# Smart IoT-Biofloc water management system using Decision regression tree

Samsil Arefin Mozumder[1*] and A S M Sharifuzzaman Sagar[2*]

[1] East Delta University, Chattogram; Bangladesh.
[2] Sejong University, Seoul, South Korea
samsilarefin313@gmail.com
sharifsagar80@sju.ac.kr

**Abstract.** The conventional fishing industry has several difficulties: water contamination, temperature instability, nutrition, area, expense, etc. In fish farming, Biofloc technology turns traditional farming into a sophisticated infrastructure that enables the utilization of leftover food by turning it into bacterial biomass. The purpose of our study is to propose an intelligent IoT Biofloc system that improves efficiency and production. This article introduced a system that gathers data from sensors, store data in the cloud, analyses it using a machine learning model such as a Decision regression tree model to predict the water condition, and provides real-time monitoring through an android app. The proposed system has achieved a satisfactory accuracy of 79% during the experiment.

## 1 Introduction

Aquaculture is proliferating across the globe these days, and an indoor fish farm known as Biofloc technology (BFT) is becoming more widespread in fish farming. Aquatic creatures are cultivated in a large tank using Biofloc technology by altering the water quality. Biofloc technology can help us save money on foodstuffs, while an intelligent aquaculture infrastructure can help us save money on the workforce. It is a cost-effective method that is helpful to the fish's health [1, 2]. Moreover, it is regarded as a water-reuse system that is effective.

In this method, the tank's water supply is restricted, and the fish are allowed to grow over an extended period without receiving any water changes. Water samples are traditionally taken and analyzed, and the results take time to get back. It's challenging to keep track of water quality indicators and change water according to impurity levels regularly. Furthermore, Bangladesh is prone to various natural disasters, such as floods and cyclones, which significantly impact aquaculture in both reservoirs and marine areas. For such catastrophes, fish producers must suffer a significant loss due to contaminated water and rising salinity in coastal waters. Conventional fish farming contributes to a number of additional difficulties, including carbon dioxide, ammonia, and nitrogen pollution in the water supply. Detoxification is a costly and time-consuming process that involves additional treatment. Biofloc is a great alternative to the conventional farming approach since it is more cost-effecient. Biofloc aids in the natural purification of water, reducing the need for additional equipment or substances. Maintaining water quality may result in increased output and decreased fish mortality. The most critical elements affecting the operation of a fish farm employing Bioflocs are the water quality characteristics.

IoT is now playing a significant role in the intelligent aquaculture industry, allowing for remote monitoring of many conditions. The Internet of Things-based fish farming is an excellent option for resolving the significant issue that occurs in Biofloc. Additionally, there is a significant economic disparity between intelligent farming and conventional farming. The Internet of Things simplifies fish farming by using a range of sensors and actuators in combination with robust communication infrastructure.

The article proposes an Internet of Things-based intelligent water monitoring system for Biofloc technology in Bangladesh. The proposed system monitor and measures water quality using sensors capable of sensing the quality of water components such as dissolved oxygen, nitrogen, pH, water temperature, nitrate, ammonium, and carbon dioxide, and transmitting the measured data to a mobile device through a mobile application. Moreover, the system automatically controls the actuators to resolve the water quality issues if the water quality level exceeds a certain threshold. Fish producers can also resolve this issue by examining the measured values of water quality parameters from their smartphones.

## 2 Related works

Researchers have suggested different water quality management systems and control methods for aquaculture; however, they concentrated on a few types of sensors such as pH, temperature, and water level, while excluding water regulating actuators. Zougmorc et al. [3] developed low-cost IoT solutions for African agricultural fish producers; however, the system lacks an automated actuator that regulates water parameters. Encinas et al. [4] presented a ZigBee-based wireless sensor network for aquaculture systems that included a temperature sensor, a pH sensor, and a Dissolved Oxygen sensor. However, water parameter management is not performed here as well. Liu et al. used the Recirculating Aquaculture System (RAS) to conduct an experiment on "Ras Carpio" [5]. In 2011, RAS was a better solution to pond-based aquaculture. The water parameters were continuously checked, and if the value of a parameter exceeded specific parameters, the water automatically recirculated. WATT TriO Matic, WATT Sensolyt, and WATT Tri oxyTherm type sensors were used to measure dissolved oxygen, pH, and temperature. While the method had many advantages and was quickly superseded by a conventional aquaculture system, it did have certain drawbacks. It requires water exchange, which is a time-consuming and expensive procedure. Dzulqornain et al. developed a novel IoT system based on the IFTTT architecture [7]. They measured dissolved oxygen levels, the temperature of the water, and the PH level. Sensors were used to determine the water level, and an aerator system was used to regulate the system. The aerator system was integrated with a microcontroller NodeMCU, a relay, a power source, and a propeller. Sensor data was transferred to the web, so the customer could see it from any location. Teja et al. [8] proposed a real-time management system for fish ponds using the ESP 32 development board, AWS cloud, and sensor technologies. This system is made up of a pH sensor, ultrasonic sensor, DHT 11 sensor for determining the quality of water. This system monitored sensors; however, the suggested system's shortcoming is that actuator control of water quality parameters was not performed.

To optimize production, individuals must examine the following water quality parameters and adhere to the ethical requirements for fish farming. Water temperature is a critical factor that has a significant impact on fish farming. Fish larvae can live at varying temperatures. Fish have a poor tolerance for temperature changes that occur rapidly. The temperature has other detrimental effects since algae and zooplankton are also temperature sensitive. It controls oxygen levels, pH, salinity, and other temperature-dependent water characteristics both directly and indirectly. As the temperature rises, dissolved oxygen levels drop, and more carbon dioxide is generated in the water due to the lack of oxygen in hot water |9| [10]. For better fish health and development, the water in the tank must be between 24 and 30 degrees Celsius [11]. However, the water temperature has an effect on the fish, as does the pH (Potential of Hydrogen) value indicating the acidity or alkalinity of a solution. The water in the tank may be acidic (pH 7.0)—alkaline (pH greater than 7.0). As the pH of the water decreases, the ammonia ions ($NH_3$) chemically react to form ammonium ions ($NH_4^+$) and hydroxyl ions ($OH^-$) [12]. And carbon dioxide ($CO_2$) As pH levels rise, they combine chemically with water to produce toxic ammonia ions ($NH_3$) that harm and kill fish [11].

## 3  Proposed System Architecture

The architecture of our Intelligent Biofloc water quality monitoring system is shown in Figure 1. The system constantly monitors and transmits alerts to a smartphone through cloud infrastructure using an ESP32 development kit module, allowing users to access real-time data from anywhere. The project's primary goal is to examine the water's pH, temperature, TDS, and electrical conductivity. We also utilized machine learning techniques such as a decision regression tree on the Biofloc dataset to train and predict the water quality of the Biofloc container. The system's sensors give a real-time measurement of the water's qualities. The system produces output by analyzing the training data and the sensor readings. It forecasts water quality, assesses the situation, and automatically controls actuators to maintain a specific level of water quality. It notifies the user of the result and option through an App, displaying real-time water metrics.

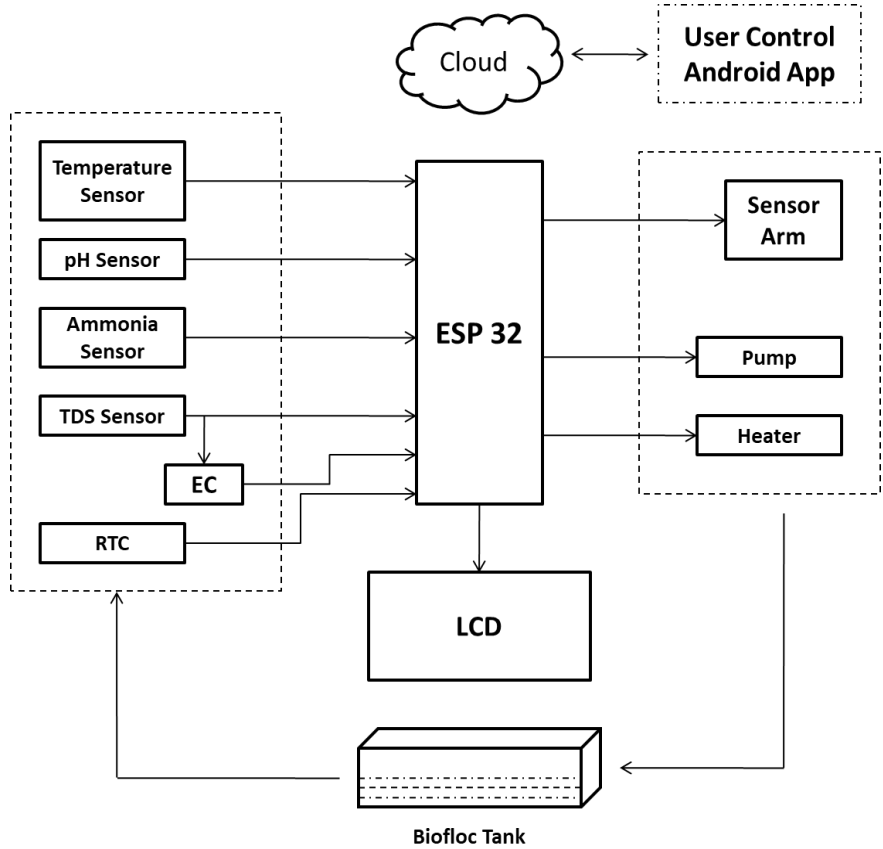

**Fig. 1.** The proposed system architecture of smart IoT Biofloc water management system

### 3.1 Hardware architecture

The hardware architecture mainly consists of three parts such as the Perception, IoT cloud, and Graphical user interface (GUI). The components used in the perception parts are PH Sensor, Temperature sensor, TDS sensor, Ammonia Sensor, RTC, LCD, ESP32 development Kit, Sensor Arm, Pump, Heater. We have used google firebase cloud service to store the real-time data. Moreover, we have developed an android application for the user to check and control the water quality remotely.

**PH sensor:** The PH sensor detects the flow of hydrogen ions in liquid systems, showing causticity or alkalinity as PH. For ESP32, we utilize a PH sensor, model PH sensor module probe. This sensor is used in our research to measure soil acidity and water quality.

**Water temperature sensor:** A water temperature sensor, model DS18B20, is used in this system. It can accurately detect temperatures ranging from 55 to 125 degrees Celsius.

**Ammonia sensor:** One of the most serious threats to fish and shrimp farming viability is ammonia. We have used an ammonia sensor to monitor ammonia of the Biofloc system. We have used the MQ-135 Gas sensor to collect ammonia data.

**TDS sensor:** The TDS value shows the number of soluble solids dissolved in one liter of water. Water with a high total dissolved solids (TDS) value is generally considered to be less clean than water with a lower TDS value. As a result, the TDS value may be utilized as one of the indicators of water quality. We have used the TDS meter to measure the turbidity of the water.

**Electrical conductivity:** Biofloc's primary objective is to provide and maintain an appropriate and stable habitat for fish. Electrical conductivity (EC) is a property of water that indicates its capacity to "transport" an electrical current and, therefore, an approximation for the concentration of dissolved particles or ions in the water. Pure water has a very low conductivity; therefore, the more dissolved particles and ions in the water, the more electrical current it can conduct. EC is critical in checking the water's quality.

**ESP32 development board:** The ESP32 development board is the successor of the ESP8266 kit. The ESP32 has several additional features over its predecessor. It has dual-core processing and wireless connectivity through Wi-Fi and Bluetooth.

**IoT cloud:** We can store all sensor data in the Biofloc water monitoring system with the help of IoT cloud service. It also provides the system's visual analysis. In our research, we have used the google firebase cloud servers to store and transfer data to the user end.

**GUI:** GUI is essential for real-time IoT monitoring systems. Users can monitor sensor data and control actuators using the GUI remotely. We have made an android app using android studio for the proposed system. The app is connected with the google cloud and acquires the stored data from the cloud. Users can check real-time data using the app. We have also made it available for the user to control the actuators through the app.

### 3.2 Control System Architecture

Algorithm 1 describes the control procedure of the proposed Biofloc water monitoring system. The proposed system uses PH, water temperature, Ammonia, TDS, and electrical conductivity to process and analyze the data. The system then predicts the water condition and control actuator based on the results of the water condition prediction. Firstly, the system saves the sensor data to the relevant variables; then it sends the data to the IoT cloud for storing and real-time monitoring. Secondly, the IoT cloud sends the data to the user android app, where users can check real-time sensor data and control the actuators manually. Thirdly, sensor data are also fed into the classifier to predict the water condition of the Biofloc system. If the water condition is not good, the system automatically checks every sensor data to control the actuators.

---

**Algorithm 1:** Smart IoT Biofloc water monitoring system

**Input** : PH sensor, Water temperature sensor, Ammonia sensor, TDS sensor, Electrical conductivity

**Output:** Water condition prediction, actuator control

1 $var \leftarrow PHsensor, Watertemperaturesensor, Ammoniasensor, TDSsensor, Electricalconductivity$
2 $var \rightarrow IoTcloud$
3 $IoTcloud \rightarrow Androidapp$
4 $var \rightarrow classifier$
5 $condition \leftarrow classifier$
6 **if** $condition \neq good$ **then**
7     **if** $temperature > 30$ **then**
8         Turn on Oxygen Pump and cooling fan
9     **if** $temperature < 24$ **then**
10        Turn on heater
11     **if** $PH > 9$ **then**
12        Turn on oxygen pump and add phosphoric acid
13     **if** $PH < 6.5$ **then**
14        Turn on water filter and add baking soda
15     **if** $turbidity > 1800 \| turbidity < 1100$ **then**
16        Turn on water filter
17 **else**
18     Turn off all actuators

---

### 3.3 System prototype

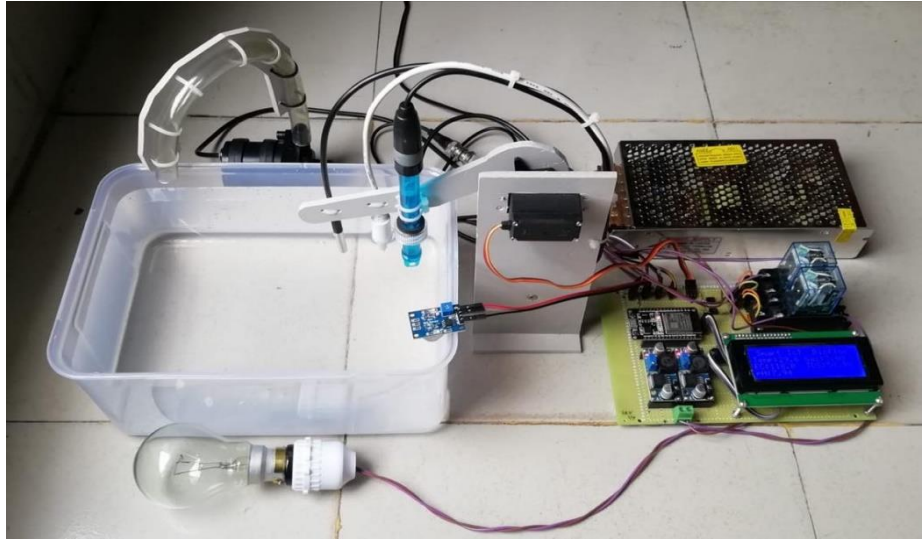

**Fig. 2.** Proposed smart IoT Biofloc water management system prototype

Figure 2 shows the proposed system prototype. We have used a water bulb to demonstrate the water heater. A small pump is used to demonstrate the oxygen pump, and a servo motor is used to demonstrate the cooling function of the proposed system. We have also attached a sensor arm equipped with a TDS and temperature sensor probe to clean the prob from time to time. The sensor arm is controlled through a servo motor where an android application can control the servo motor.

Figure 3 shows the central control and sensing board of the prototype. The control board consists of a microcontroller, relay, and an LCD panel to show the real-time sensor data. Sensors and actuators are attached to the ESP32 microcontroller through the PCB board.

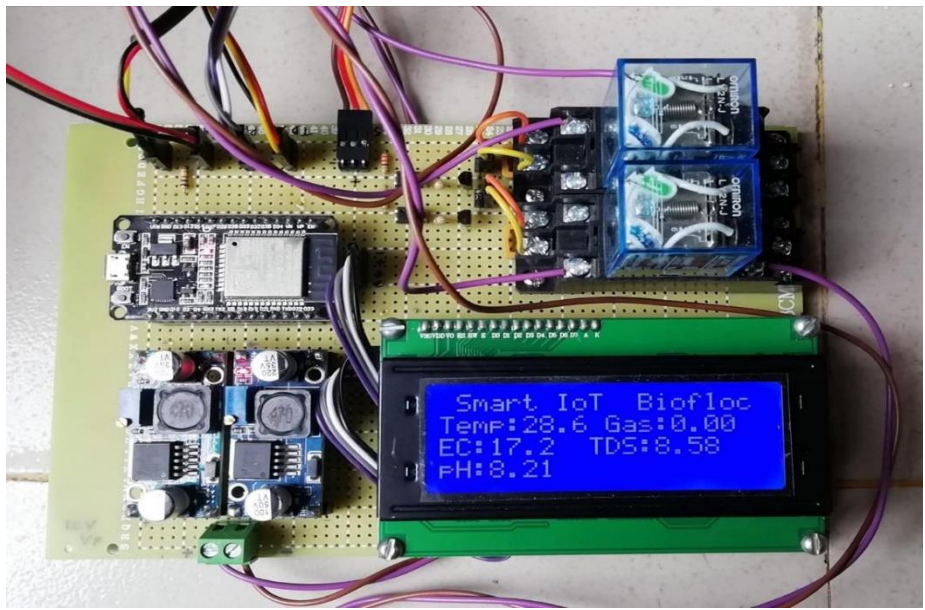

**Fig. 3.** The proposed central control unit for sensors and actuators

Figure 4 shows the raw sensor data stored in the google firebase cloud from the prototype. Real-time sensor data with actuators conditions are stored and sent to the user android application through the IoT cloud. Users can access this data from anywhere in the world.

Figure 5 the raw sensor data collected from the prototype in our developed android application. The app is connected to the IoT cloud, and it updates the sensor data in real-time. Temp represents the temperature, PH represents the PH lev-el, EC represents the electrical conductivity, Gas represents the Ammonia gas level, and turbidity represents the water's turbidity level. We have also added three switches to control the actuators remotely in an emergency if the automatic system fails to work.

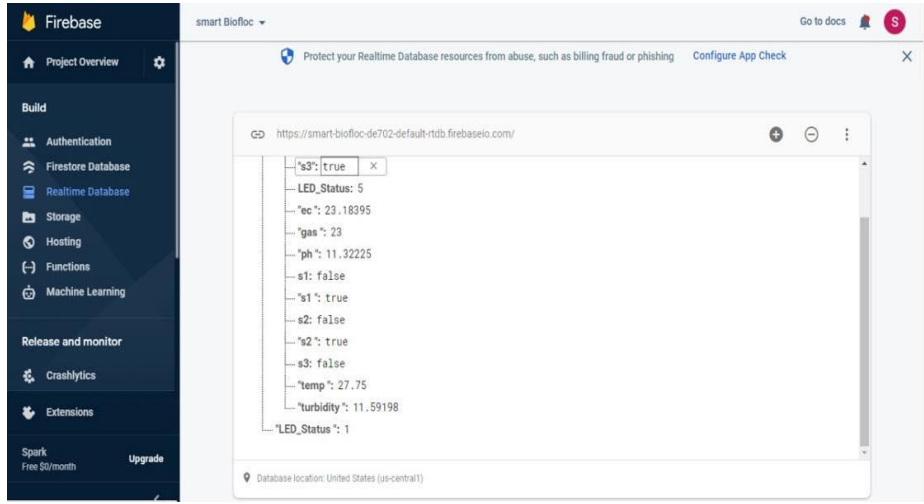

**Fig. 4.** Sensor data stored in the Google Firebase IoT cloud

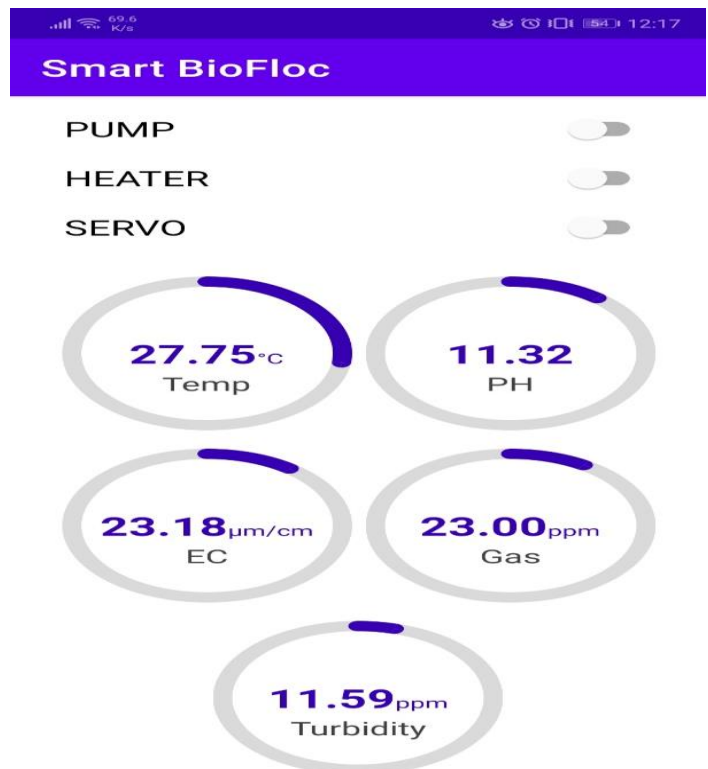

.

**Fig. 5.** The GUI of the developed Smart IoT Biofloc water management system android application.

### 3.4 Machine Learning Model

The decision regression tree machine learning model is used in the proposed system to predict the water condition of the Biofloc container. The decision tree model performs very well when dealing with tabular data with numerical characteristics or reference variables with less than hundreds of categories. Decision trees, as opposed to linear models, may capture non-linear interactions between characteristics and targets.

We have taken 4000 samples using the temperature, PH, and TDS sensor to train our classifier. The data was processed and stored into a specific string format to extract the features from the sensor data. Many python libraries provide decision regression trees to train datasets. We have used Microml python library to train the model because it also provides the porting facility to the microcontroller. Then we have used the EloquentArduino library in the Arduino IDE for ESP32 to make use of the ported classifier to predict real-time water conditions.

## 4 Result Analysis

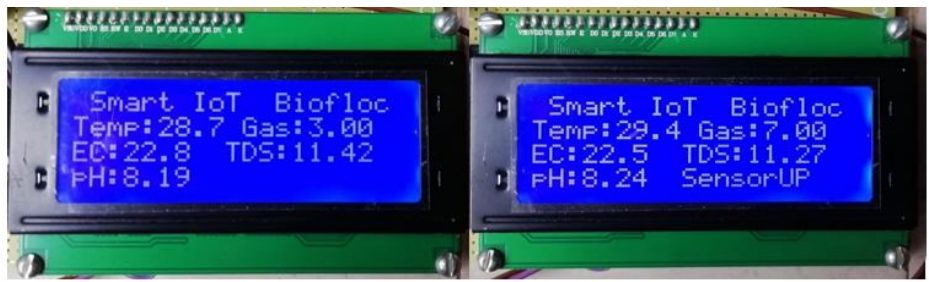

**Fig. 6.** Sensor value displayed in the LCD panel during the experiments.

The proposed system is successfully implemented in the Biofloc system. We have observed simultaneous data transmission from the system to the cloud and android app. Figure 6 shows the real-time sensor data in the LCD panel during the experiment. It is easily observable that our system successfully collects the sensor data. As water condition is very important for healthy fish production, actual sensing data will keep the water condition at a suitable level to accelerate the fish production without any need of humans.

We have prepared the dataset of 4000 samples for regression tree classification training. We have used temperature, PH, Turbidity as key parameters to determine the water condition of the Biofloc system. We have also set a certain threshold for the key parameters for the good condition, such as temperature (24-30), PH (6-9), and Turbidity (1200-1800). If any of the corresponding values of the parameters go beyond the threshold, we label the data as bad water condition. Table 1 shows the 10 samples from the prepared dataset used to train the classifier. These data were fed into the regression tree machine learning model for training. We have used the Micromlgen python library to train the data. Micromlgen provides different machine learning models to train and port the results in c++ files for use in the microcontrollers. The generated c++ file can be used as a classifier library in the microcontroller for the water condition prediction.

**Table 1.** Dataset samples used in the machine learning model

| Condition | Temperature (°C) | PH | TDS |
|---|---|---|---|
| Good | 24.98 | 7.81 | 1350 |
| Good | 25.59 | 7.81 | 1760 |
| Good | 26.03 | 7.98 | 1750 |
| Good | 25.95 | 7.95 | 1740 |
| Good | 26.31 | 7.66 | 1740 |
| Bad | 23.00 | 9.90 | 1850 |
| Bad | 31.00 | 7.00 | 1400 |
| Bad | 23.00 | 9.00 | 1600 |
| Bad | 33.00 | 5.70 | 1050 |
| Bad | 27.90 | 7.16 | 1100 |

Table 2 shows the different sensor values collected from the Biofloc system for a day. We can see that early morning and the evening temperature of the water was very low. During the midday, the temperature was higher than in the morning, which also affects the other parameters of the Biofloc system, such as PH, EC, turbidity. The gas level was stable throughout the day.

**Table 2.** Sensor output during the experiment in the Biofloc system

| Water Quality Parameter | | | | | |
|---|---|---|---|---|---|
| Time | Temperature (°C) | pH | TDS (Mg/L) | EC μm/cm | NH3 ppm |
| 5:00 AM | 25.56 | 8.1 | 1752 | 45.85 | 5.95 |
| 5:30 AM | 25.59 | 7.98 | 1760 | 45.65 | 5.96 |
| 6:00 AM | 26.03 | 7.98 | 1750 | 45.65 | 5.98 |
| 6:30 AM | 25.95 | 7.95 | 1740 | 45.62 | 5.98 |
| 7:00 AM | 26.31 | 7.66 | 1740 | 45.63 | 5.65 |
| 7:30 AM | 27.45 | 7.54 | 1740 | 45.64 | 5.89 |
| 8:00 AM | 27.32 | 7.36 | 1741 | 45.58 | 5.91 |
| 8:30 AM | 27.35 | 7.20 | 1739 | 45.59 | 5.91 |
| 9:00 AM | 27.86 | 7.19 | 1734 | 46.02 | 5.97 |
| 9:30 AM | 27.90 | 7.16 | 1734 | 45.99 | 6.07 |
| 10:00AM | 28.21 | 7.18 | 1732 | 46.00 | 6.09 |
| 10:30AM | 28.25 | 7.17 | 1730 | 46.10 | 6.08 |
| 11:00AM | 28.43 | 7.16 | 1728 | 46.11 | 6.11 |
| 11:30AM | 29.06 | 7.16 | 1735 | 46.15 | 6.13 |
| 12:00PM | 29.12 | 7.16 | 1729 | 46.16 | 6.11 |
| 12:30PM | 29.45 | 7.17 | 1728 | 46.17 | 6.13 |
| 1:00 PM | 29.33 | 7.14 | 1726 | 46.22 | 6.26 |

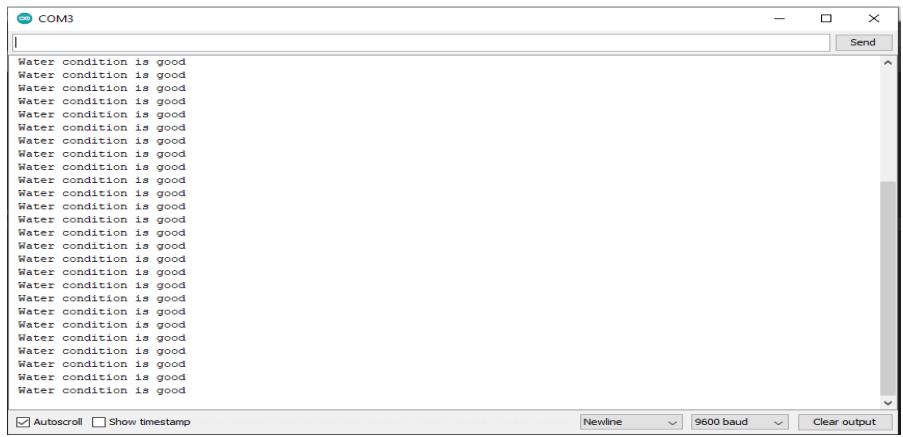

**Fig. 7.** Predicted water condition level in the ESP32 serial monitor.

Figure 7 shows the predicted water condition in the Arduino IDE. The classifier takes 500ms to predict the water condition from the acquired sensor data of the Biofloc system. We can see that the water condition of the Biofloc was good during the experiment.

Previous research works on smart Biofloc systems are based on Long Short-Term Memory (LSTM) [14], Least-Squares Support Vector Regression (LSSVR) [15], Group Data Handling Method (GMDH) [16], Support Vector Machine (SVM) [17], and Artificial Neural Network (ANN) [18-19] which measure pH, DO, BOD, COD, TDS, EC, $PO_4^{3-}$, $NO_3$-N, and $NH_3$-N, and continuously compare with the acceptable range.

Some research proposed manual control over the app; most of them could take the reading from the sensors by a microcontroller and display them on LCD. Those systems always had to have a person to monitor and manually manipulate parameters accordingly. Our proposed system can predict water conditions and maintain the whole process in a precise way. It needs only a small portion of space as compared to the pond and wide swampland. A container is enough to produce a large number of fish in the house.

We conducted a comparison of our implemented system to pre-existing systems. We highlighted the method's precision, parameters, prediction, and decision-making capabilities. The comparison is shown in table 3. We can see that our approach achieved a good accuracy of 79%, which outperforms some existing methods. Moreover, we have also monitored different parameters that were not included in the existing system, such as ammonia gas monitoring. However, the LSTM method [14] yields good accuracy, but they did not provide any automatic control system for water monitoring, which is very important for IoT-based systems. Therefore, Our proposed system can be considered a robust Biofloc water condition monitoring system.

Table 3. Performance comparison between different proposed water monitoring systems

| Method | Accuracy | Parameters | Solution | Smart-Decision |
|---|---|---|---|---|
| **LSTM [14]** | 82% | pH, temperature, DO | Manual control | Yes |
| **LSSVR [15]** | 76% | Temperature, DO, TDS, pH, EC, PO43-, NO3-N | Manual control | Yes |
| **GMDH [16]** | 74% | Temperature, DO, TDS, pH | Manual control | Yes |
| **SVM [17]** | 70% | Temperature, DO, TDS, pH | Manual control | Yes |
| **ANN [18]** | 72 % | Temperature, DO, TDS, pH, BOD, COD, TSS | Manual control | Yes |
| **ANN [19]** | 77 % | Temperature, DO, TDS, pH, Floc | Automatic control | Yes |
| **Proposed method** | 79% | Temperature, DO, TDS, EC, pH, $NH_3$ | Automatic control | Yes |

## 5 Conclusion

Fish farming has been thriving for decades despite economic limitations, increasing prices or even non-availability of personnel, regular water management, and rapid increases in pollution. The smart system analyses water conditions in real-time, lowering production costs, increasing productivity, reducing human dependence, and ensuring socio-economic sustainability. The suggested system checks water quality in real-time and provides an immediate warning to the user. A machine learning method such as a decision regression tree was used to maintain a sustainable water condition. Experiments on the implemented features were conducted to validate the proposed IoT system. According to the tests, 79% accuracy was achieved. In the future, we'd want to enhance the model's accuracy and assess how well it performs in terms of the fish growth in the Biofloc.

## References

1. B. Ghose, "Fisheries and Aquaculture in Bangladesh: Challenges and Opportunities.", Aquaculture Research, 2014. Bashar, Abul. (2018).Biofloc Aquaculture: prospects and challenges in Bangladesh. 10.13140/RG.2.2.13233.94560.

2. D. SK, "Biofloc Technology (BFT): An Effective Tool for Remediation of Environmental Issues and Cost Effective Novel Technology in Aquaculture", International Journal of Oceanography & Aquaculture,vol. 2, no. 2, 2018. Available: 10.23880/ijoac-16000135.
3. T.-W. Zougmore, S. Malo. F. Kagembega, atul A. Toguevini. "Low cost IoT solutions for agricultures fish fanners in Afirca:a case study from Burkina Faso." in 2018 1st International Conference on Smart Cities and Communities (SCCIC). Ouagadougou, Jul. 2018. pp. 1-7.
4. C. Encinas. E. Ruiz. J. Cortez, and A. Espinoza, "Design and implementation of a distributed IoT system for the monitoring of water quality in aquaculture," in 2017 Wireless Telecommunications Symposium (WTS). Chicago, IL, USA, Apr. 2017, pp. 1-7.
5. S. Liu et al., "Prediction of dissolved oxygen content in river crab culture based on least squares support vector regression optimized by improved particle swarm optimization", Computers and Electronics in Agriculture, vol. 95, pp. 82-91, 2013. Available: 10.1016/j.compag.2013.03.009
6. Author, F., Author, S.: Title of a proceedings paper. In: Editor, F., Editor, S. (eds.) CONFERENCE 2016, LNCS, vol. 9999, pp. 1–13. Springer, Heidelberg (2016).
7. Dzulqornain, M.I., Rasyid, M.U., & Sukaridhoto, S. (2018). Design and Development of Smart Aquaculture System Based on IFTTT Model and Cloud Integration.
8. K. B. R. Teja. M. Monika. C. Chandravathi. and P. Kodali, "Smart Monitoring System for Pond Management and Automation in Aquaculture." in 2020 International Conference on Communication and Signal Processing (ICCSP). Chennai. India. Jul. 2020. pp. 204-208
9. M. F. Saaid. N. S. M. Fadhil, M. S. A. M. Ali. and M. Z. H. Noor. "Automated indoor Aquaponic cultivation technique," in 2013 IEEE 3rd International Conference oil System Engineering and Technology, Aug. 2013. pp. 285-289.
10. Claude E. Boyd, "Water temperature in aquaculture « Global Aquaculture Advocate," Global Aquaculture Alliance,
    https: Www.aquaculturcalhance.org advocate water-temperature-in aquaculture.
11. A. Bhatnagar and P. Devi. "Water quality guidelines for the management of [Kind fish culture," vol. 3, p. 30, 2013.
12. Tucker. Craig S.. and Louis R. D'Abramo. Managing high pi I in freshwater ponds. Stoncvillc: Southern Regional Aquaculture Center, 2008.
13. F. Kubitza, "The oft-overlooked water quality parameter:pH « Global Aquaculture Adv cate," Global Aquaculture Alliance. https://www.aquaculturealliance.oig/advocaterihe-of overlookedwater-quality-parameter-ph.
14. Hu, Z., Zhang, Y., Zhao, Y., Xie, M., Zhong, J., Tu, Z., & Liu, J. (2019). A water quality prediction method based on the DEEP LSTM network considering correlation in Smart Mariculture. *Sensors*, *19*(6), 1420. https://doi.org/10.3390/s19061420
15. Tan, G., Yan, J., Gao, C., & Yang, S. (2012). Prediction of water quality time series data based on least squares support vector machine. *Procedia Engineering*, *31*, 1194–1199. https://doi.org/10.1016/j.proeng.2012.01.1162
16. O. Varis, "Water Quality Models: Tools For The Analysis Of Data, Knowledge, and Decisions", Water Science and Technology, vol. 30, no. 2, pp. 13-19, 1994. Available: 10.2166/wst.1994.0024.
17. Haghiabi, A. H., Nasrolahi, A. H., & Parsaie, A. (2018). Water quality prediction using machine learning methods. *Water Quality Research Journal*, *53*(1), 3–13. https://doi.org/10.2166/wqrj.2018.025
18. Kadam, A. K., Wagh, V. M., Muley, A. A., Umrikar, B. N., & Sankhua, R. N. (2019). Prediction of water quality index using artificial neural network and multiple linear regression modelling approach in Shivganga River Basin, India. *Modeling Earth Systems and Environment*, *5*(3), 951–962. https://doi.org/10.1007/s40808-019-00581-3

19. Rashid, M.M., Nayan, A., Simi, S.A., Saha, J., Rahman, M.O., & Kibria, M.G. (2021). IoT based Smart Water Quality Prediction for Biofloc Aquaculture.